\documentclass[conference]{IEEEtran}
\IEEEoverridecommandlockouts
\usepackage{cite}
\usepackage{amsmath,amssymb,amsfonts}
\usepackage{algorithmic}
\usepackage{graphicx}
\usepackage{textcomp}
\usepackage{xcolor}
\def\BibTeX{{\rm B\kern-.05em{\sc i\kern-.025em b}\kern-.08em
    T\kern-.1667em\lower.7ex\hbox{E}\kern-.125emX}}
\begin{document}

\title{Why Data Science Projects Fail\\
{\footnotesize Understanding of Data Science Project failure and reasoning}
}

\author{\IEEEauthorblockN{Balaram Panda}
\IEEEauthorblockA{\textit{Software Engineering} \\
\textit{University of Auckland}\\
Auckland, New Zealand \\
bpan575@aucklanduni.ac.nz}

}

\maketitle

\begin{abstract}
Data Science is a modern Data Intelligence practice, which is the core of many businesses and helps businesses build smart strategies around to deal with businesses challenges more efficiently. Data Science practice also helps in automating business processes using the algorithm, and it has several other benefits, which also deliver in a non-profitable framework. In regards to data science, three key components primarily influence the effective outcome of a data science project. Those are 
\begin{enumerate}
    \item Availability of Data
    \item Algorithm
    \item Processing power / infrastructure 
\end{enumerate}

In today's technology world, there is no limitation on data as well as processing power, and we have a much more efficient algorithm to produce the desired output. In spite of the success of Data Science projects, many Data Science projects still fail and are unable to produce the desired outcome. In this paper, we have explored the bottleneck of Data Science projects and provided some recommendations to make data science projects more successful. 
Standard Data Science project development life-cycle CRISP-DM \cite{b1} is old in this agile development\cite{b31} world, and but most Data Science practices still follow CRISP-DM. In general, Data Scientist analyses scenarios where a predictive model or machine learning model might fail. But this study is to analyze when and why the Data Science project fails despite an excellent model. Data Science is a diverse field. It needs technical as well as business knowledge to deliver a project. Hence, to understand why the Data Science project fails, we need to understand challenges from the business side and technical side.
\begin{enumerate}
    \item Technical perspective
    \item Business leader perspective or stakeholder perspective
\end{enumerate}

Also, it has been observed that the success of the Data science project depends on business domain. Example market propensity model, a strategic market campaign method, is more successful in any retail use case than a fraud analytic model in the banking fraud domain. So domain agnostic framework was implemented in this research to make this research independent of the business domain.

\end{abstract}


\section{Introduction} 

Data Science is comparable to applied research. Often in research, there are lots of repetitive processes to be followed until it generates the desired outcome. Similarly, data science processes are often repetitive until they meet the business objectives. Those business objectives are set during the repeated Business Understanding and Data Understanding phase of CRISP-DM. 
CRISP-DM developed during 90’s, and since then, components of Data Science have evolved a lot in regards to the way data is collected, volume, velocity, and variety of the information being analyzed(big data \cite{b24}), processing methodology, algorithms, and outcome delivery. Hence it is crucial to understand in detail the current day's Data Science challenges, deep dive into the reasoning of failures, and find out the best practice. Data Science is an evolving field, and lots have been changed since last decade in regards to various keys areas, such as data collection, big data processing on cloud and/or parallel or distributed processing framework, evolution of new algorithm, data visualization, deployment and productionizing Data Science solutions as a digital product etc. There are several challenges in those key areas; as an example, because of big data, the Data Science project collects lots of irrelevant and uncleaned data. Sometimes the intermediate data source doesn’t have the correct business logic, which is mapped to the business objective of the data science project.  “big data brings lots of “big errors” in data quality and data usage, which cannot be used as a substitute for sound research design and solid theories.” \cite{b4}  Data quality is one of the common challenges in the Data Science Project. Another major challenge is domain understanding, and many times domain understanding is a bottleneck for Data Science project success. Also, in many cases, data science issues are seen differently and vary from one domain to another, stakeholder view and data scientist view. As mentioned by Dr.Om Deshmukh \cite{b9} " Data Science project success also depends on what stakeholder knows about the expected outcome of the data science project and how effectively stakeholder expectations being managed though effectivthroughunication." 
   Data Science projects struggles with these types of challenges and more other challenges in industry settings, but not enough research has been done in this space to highlight those problems and their possible solutions.  For better quality research in this domain, years of experience in the Data Science field and role/designation level diversity in data science projects such as key stakeholders and data scientists have been included. 
   

We prepared a list of questions and interviewed experienced Data Scientists, Senior Leader who actively manage data science and data analytic team, and stakeholders part of data science projects, in a motivation to understand the perception on data science project failure. Then did the thematic analysis to analyze and draw conclusions from data. Then came up with the most popular theme by thoroughly analyzing the discussion. 

In this research, we have discovered and highlighted the key challenging areas where Data Science practice and Data scientists should focus. Also, some suggestions on tackling those challenges. Also, we came up with a state of art framework to tackle those challenges in efficient ways.

\section{Research Methodology}

\subsection{Research methodology}
In this research Interview based semi-structured Qualitative research methodology \cite{b8} have been used. 
\subsection{Interview Design}
In this research the objective is also to capture the diversity in regards to the people involved in a data science project such as Data Scientist and business owner and/or data science capability leader and/or stakeholder. Hence for better outcome, two types of interview questions have prepared
\begin{enumerate}
    \item[]
    \begin{enumerate}
    \item[]
    \begin{enumerate}
     \item[\textbf{Group A:}] Data Scientist with some years of experience working in Data Science project
     \item[\textbf{Group B:}] Business owner / leader / stakeholder
     \end{enumerate}
    \end{enumerate}
\end{enumerate}

The interviews had done via zoom software.
During the interview, answers and key points were captured. Also, interview voice recording has been stored in google drive for reference.

\subsection{Candidate selection for interview and Sample Diversity}
Interview candidates contacted via reference and Linkedin(professional networking website). Manually reviewed their background and put them into the Group A or Group B Category based on their experience.
Some candidates were also selected through reference. While selecting a reference candidate, Snowball technique \cite{b5} had been used to select the reference candidate. Snowball helped to capture unbiased interviews candidate samples.
During the selection process, candidates domain experience deiversity such as Banking and finance, Retail, Software Product, IT Services, Telecommunication etc. had been considered. In this research the objective also is to add domain diversity so that Data Science problems can be viewed from a normalized and generic approach. Also to capture more diversity candidate selected from various geography such as India, USA, Canada in addition to New Zeland. 

\begin{table}[htbp]
\caption{List of Interview Candidates}
\begin{tabular}{|l|l|l|l|l|}
\textbf{ID} & \textbf{\begin{tabular}[c]{@{}l@{}}Year Of \\ Experience\end{tabular}} & \textbf{\begin{tabular}[c]{@{}l@{}}Group A or \\ Group B\end{tabular}} & \textbf{\begin{tabular}[c]{@{}l@{}}Business \\ Domain\end{tabular}}  & \textbf{\begin{tabular}[c]{@{}l@{}}Working \\ Geography\end{tabular}} \\
\hline
\hline
C1                    & 5 +                         & Group A                                                                & \begin{tabular}[c]{@{}l@{}}IT and \\ Consulting\end{tabular}         & USA                                                                   \\
\hline
C2                    & 10 +                        & Group B                                                                & \begin{tabular}[c]{@{}l@{}}Banking and \\ Finance\end{tabular}       & NZ                                                                    \\
\hline
C3                    & 10 +                        & Group A                                                                & \begin{tabular}[c]{@{}l@{}}Software \\ Product\end{tabular}          & India                                                                 \\
\hline
C4                    & 1 +                         & Group A                                                                & Research                                                             & NZ                                                                    \\
\hline
C5                    & 5 +                         & Group B                                                                & \begin{tabular}[c]{@{}l@{}}Software\\ Product\end{tabular}           & Canada                                                                \\
\hline
C6                    & 10 +                        & Group B                                                                & \begin{tabular}[c]{@{}l@{}}IT Service and \\ Consulting\end{tabular} & India                                                                 \\
\hline
C7                    & 10 +                        & Group B                                                                & \begin{tabular}[c]{@{}l@{}}Banking and \\ Finance\end{tabular}       & NZ                                                                    \\
\hline
C8                    & 5 +                         & Group A                                                                & \begin{tabular}[c]{@{}l@{}}Software\\ Product\end{tabular}           & USA                                                                   \\
\hline
C9                    & 10 +                        & Group B                                                                & Consulting                                                           & NZ
                                                             \\
\hline                                                           
\end{tabular}
\end{table}

\subsection{Preparation of research Questions}
In first iteration a list of questions has been prepared from prior knowledge and after going through literature review papers \cite{b10} \cite{b11} \cite{b12}
Then questions are discussed with the research supervisor. Based on the discussion few questions are added, and a few are modified. A final list of questions has been prepared and shared with the supervisor in the second iteration. Then after 2nd interview, 2 more questions were added to make the interview more target-oriented. 

\begin{enumerate}
    \item[]
    \begin{enumerate}
    \item[]
    \begin{enumerate}
     \item[\textbf{Group A:}] Data Scientist with some years of experience working in Data Science project
     \item[\textbf{Group B:}] Business owner / leader / stakeholder
     \end{enumerate}
    \end{enumerate}
\end{enumerate}

\subsection{Research Questions}

\begin{enumerate}[\textbf{Group A Questions}]

\item What are the key reasons for a Data Science project success as per your experience
\item What are key reasons for Data Science project failure as per your experience
\item What steps Data Science should follow to match user expectations? Can you refer to any specific project and explain it in detail?
\item How stakeholder can contribute towards Data Science project success, and how important is it?
\item How value measured in the Data Science project and who are involved and what are the challenges? Any example based on a specific project?
\item Do you think infrastructure is a challenge for any data science project?

\item How important Data Quality is for Data Science projects?
\item What are the hindrances to maintaining data quality?
\item What are the solutions to overcome those hindrances?
\item What testing approaches do you take to test the model based on business user expectations?
\item Some detailed discussion on failure mentioned in open ended. Like how to solve model explainability? Is there any approach you followed, or can you suggest from your experience how to maintain model explainability and without compromising the model accuracy?
\item Does the level of competencies affect project outcome and how to solve that?
\item How often do Data Science projects productionize and the pitfalls you experienced during productionization ?
\item Many Data Science projects help automate the existing process; what are the challenges you face during knowledge elicitation from business users in those cases?
\item How challenging is change management in Data Science? 
\item There are limitations of Data usability due to governance restriction? How often do you see this as a limitation, and how do you solve these?
\end{enumerate}

In Group A, questions from 1 to 6 are open ended questions and 6 to 16 are focused questions. 

\begin{enumerate} [\textbf{Group B Questions}]

\item What are the essential things Data Scientists must do to make a Data Science project successful?
\item What is your view on Stakeholder's contribution to a Data Science project?
\item What are the critical gaps between a successful and unsuccessful Data Science Project from stakeholder/business owner's perspective?
\item What improvement do you want to make towards the traditional Data Science approach to make it more valuable for business?
\item How important change management is for a Data Science Project? Any suggestions to make change management successful?
\item Data Science projects have lots of intangible benefits; how do you create value metrics for Data Science projects? How often are you able to achieve that?
\item Do you follow any ROI metrics for Data Science projects?  (question 6 extension)
\item How do you define a Data Science project different from a Business Intelligence project?   (Optional question)

\end{enumerate}

\subsection{Interview Process and Data Collection}
Interviewer contacted via Linked-in and via emails. Once they confirmed their participation, the consent document was shared, and formal consent approval was taken before the interview. Interviews are done via Zoom and in-person meetings. 
During the interview, answers and critical points has been captured. Also, interview voice recording has been stored in google drive for reference. Only the question-answer part of the interview has been recorded. In an average interview, only the question-answer part lasted for 30 minutes. For the introduction, including research motivation, we spent around 8 to 10 minutes in each interview.  

\subsection{Data Understanding and Analysis}
Thematic analysis\cite{b6} has been used to analysis the data. Thematic analytic is a popular method for qualitative research. 
As suggested by Braun and Clark \cite{b30} there are 6 steps of thematic analyses as follows
\begin{enumerate}
    \item[]
    \begin{enumerate}
    \item[]
    \begin{enumerate}
     \item[\textbf{Step 1:}] Become familiar with the data
     \item[\textbf{Step 2:}] Generate initial codes
     \item[\textbf{Step 3:}] Search for themes
     \item[\textbf{Step 4:}] Review themes
     \item[\textbf{Step 5:}] Define themes
     \item[\textbf{Step 6:}] Write-up
     \end{enumerate}
    \end{enumerate}
\end{enumerate}

As the questions were designed before the interviews and interview, we familiarised ourselves with the data during the interview. Also, note has been taken during the interview. The code generation to theme generation is described in the following Theme Generation section and the write-up step mentioned in the result section.  


\subsection{Theme Generation}
There are various orientations in thematic analytic mentioned in \cite{b26,b27}. In this research, we have used a combination of the Inductive and Deductive approaches. We have prior knowledge from experience and literature review; based on that, questions have been prepared on the most common issues that have been known so far. After the interview, get to know more new issues that the Data Science project is experiencing and solutions around how to tackle that.
After the interview, we had iterate over the recordings, list down all the key issues that experts mentioned in the interview, and mapped that to the questions. Then cluster those issues together to identify the common themes in those issues. In addition to the theme, we also got solutions and workaround to tackle those issues. 

\section{Results}
\begin{table}[htbp]
\caption{Resulted Themes}
\begin{tabular}{|l|l|l|}
\textbf{Themes}                                                                                            & \textbf{\begin{tabular}[c]{@{}l@{}}Questions \\ where themes \\ were found\end{tabular}} & \textbf{\begin{tabular}[c]{@{}l@{}}Key concerns \\ for \\ Data Sciecne\\ Project Success\end{tabular}} \\
\hline     
\hline     
\begin{tabular}[c]{@{}l@{}}Effective Stakeholder\\ Managment\end{tabular}                                  & AQ4,AQ10,BQ1,BQ2                                                                             & Yes                                                                                                    \\
\hline     
\begin{tabular}[c]{@{}l@{}}Clarity in Business \\ problem understanding\end{tabular}                       & AQ1,BQ6,AQ2                                                                                     & Yes                                                                                                    \\
\hline     
Data quality Issues                                                                                        & AQ7, AQ2, AQ9                                                                            & Yes                                                                                                    \\
\hline     
\begin{tabular}[c]{@{}l@{}}Model deployment \\ and productionzaiton\end{tabular}                           & AQ3,AQ6,AQ13,BQ3                                                                                 & Yes                                                                                                    \\
\hline     
\begin{tabular}[c]{@{}l@{}}Department level \\ intigration while \\ productionizing the model\end{tabular} & BQ3,BQ5                                                                                      & Yes                                                                                                    \\
\hline     
Change Managment                                                                                           & AQ15, BQ5                                                                                & No                                                                                                     \\
\hline     
Data governance road blocker                                                                               & AQ17                                                                                     & No                                                                                                   \\
\hline

\end{tabular}
\end{table}

\subsection{Effective Stakeholder Management:} 
Stakeholders are the people representing a department or owning a business product or a group of people who benefit from a project's success. In Data Science, Stakeholder can be business process owner or product owner or end-user who is going to use the Data science outcome or product. 
Many participants mentioned how crucial it is to communicate and manage Stakeholder effectively. C1 also mentioned that data scientists need to communicate well across different project stakeholders like project managers, business stakeholders, and data engineers. "Stakeholders should share equal responsibility as data scientist to make the project successful" - C1. But often, stakeholders may not be available all the time as they have other priorities. 
Also its often found the there is a gap what Stakeholder thinks about Data Science can solve vs what data science can actually solve for business.

C1 mentioned that from her experience, she found that data science projects are more likely to successful where data scientist and stake holder work together towards a common business goal. C2 mentioned that Stakeholder should take equal responsibilities as data scientist to make the Data Science project successful. 

\cite{b13} Data science teams often struggle to use their model into business processes. Also stakeholders often can’t articulate the problems they are aiming to solve. \cite{b13} there is a significant gap between stakeholder and data science teams that needs to be recognized and addressed. \cite{b13} suggest data science need to bridge the gap between an organizational structure and leadership commitment to develop better communication, processes, and gain trust among all stakeholders.

\subsection{Business Problem Understanding:} 

All the participants mentioned that Business problem understanding is the most critical step in a data science project. Business problem understanding with effective stakeholder communication improves the success criteria of a data science project. During the business problem understanding process, the data science team gets an overview of the business process and tries to articulate the problem that the business is trying to solve. It helps the data science team get more clarity around the business objective, business process, and some understanding of data. Then Data Science team prepares the plan to check for data availability. This also helps define the target variable or dependent variable in the case of a supervised modeling approach. In this process Data Science team understands the business problem well; more accurately and define the business objective and target/dependent/outcome variable for the predictive model or another type of modeling task. More accurately, the outcome variable is defined, more chances of success of the data science project.

\subsection{Data Quality issues:} 
Data quality issue another common issue which came out as a theme.
Most participants from group a mentioned about Data Quality issue. 
Also they mentioned most teams don't want to spend much time on fixing data quality issues and focusing more on deliverable, which can be measured.  Because the time spend is not mapped to any tangible benefit at this stage. So the idea is to create something which can be used for multiple projects. Example the DW(Data Warehousing) and Cube concept got more attention when BI (Business Intelligence) generate value for organisation and BI/ DW team started creating something which can be reused for many different projects.

\subsection{Model Deployment and Production:} 

Several participants from group A and group B, described that many Data Science project is not deployable or failed due to complexity in productionzing the model or the outcome. Example an antivirus software company built a predictive model build with 99.1\% accuracy on test data. No doubt that is a that is one of the best model, but if the expectation is model should be integrated into a real-time environment, that might become tricky. That requite a software engineering skill, also the key thing is to how a model file can be integrated in the existing programming environment. These things need to be plan ahead before the project starts during the business understanding and planning phase. Based on the complexity of implementation project kick-off. If its is feasible to implement, then the production implementation plan need to be discussed and agreed with the stakeholder at the beginning of the project.

\subsection{Department level Collaboration:} 
Several group B candidates highlighted that cross department collaboration is one of the major challenges while delivering the Data Science outcome. Many times model outperform and implementation done but due to lack in department level integration project may not be useful for the organization. Example a market propensity model can produce most accurate target customer, but that need to integrated in the marketing channel. Even though its integrated, the forntline channel might de-prioritize the request due to some other priority they are having. These situation are out of control of a Data Science and Stakeholder.  

\subsection{Change Management:} 
Most of the data science project are either used for strategy or time saving or cost saving using automation to reduce manual effort. Sometimes this creates resistance from the people affected. 
Several candidates from group B mentioned that Change Management is one of the challenge in any data science implementation. But this is not the key challenge as there are strategy to tackle it.
Also few participants mentioned that the solution for the change management sometimes out of control of Data Science team. As suggested by C2 "Change management specialized area and need specialized people to handle such things." 

"Change management is not a key challenge" -C3. May be because he is working in google where automation using algorithm is always promoted.

To handle change management, "DS team need to engage with the team going to affect very early and try to gain confidence from the team affected by illustrating an explaining how this is going to improve their work" - C1. 

\subsection{Data Governance:}
Data governance processes, roles, policies and standard are getting more stronger after GDPR(General Data Protection Regulation)\cite{b21}  law. 
"Around 30\% of data science projects unable to go to production due to Data Governance rules" C3. Same issues due to data governance also mentioned by various other participants from group A and Group B.
Data governance is a must have standard for any organization using data as a asset. But to avoid data science project late failure, data governance team engagement need to be discussed at the early stage of the project. So that Data governance team can advise on alternate solution to get the data, or they can advise not to opt for this project due to the data used in Data Science project are not meeting the governance standard. 
In addition to that organisation need to adopt PII (personal identifiable information) masking approaches to mask the data.

\section{Discussion and Recommendations}
There are the top 7 themes generated [Table II] in this research. More emphasis was given to Effective Stakeholder management, Most of the interview participants mentioned the challenges during stakeholder management, and few mentioned that stakeholder management is not that challenging. Also, business leaders who sometimes act as stakeholders (in our interview panel) also agreed that there is a gap in understanding the data science process from a stakeholder lens.

Usually, Data science projects have more than one stakeholder. In this multi-stakeholder environment, it's always good to have a strategy around stakeholder management. The stakeholder can have different types of posture \cite{b15} based on the RDAP(reactive, defensive, accommodative, or proactive) strategy (table Fig 2). The same can be applied to Data Science projects. As C2 mentioned, stakeholders should share equal responsibility. Hence for a successful data science project, the strategy should be Accommodative or Proactive where stakeholder postures are Accept responsibility and Anticipate responsibility, respectively, as described in fig1. But to do that as suggested by few candidated, every Data Science project should have top management buying to make the DS project successful.

\begin{figure}[htbp]
\includegraphics[width=0.50\textwidth]{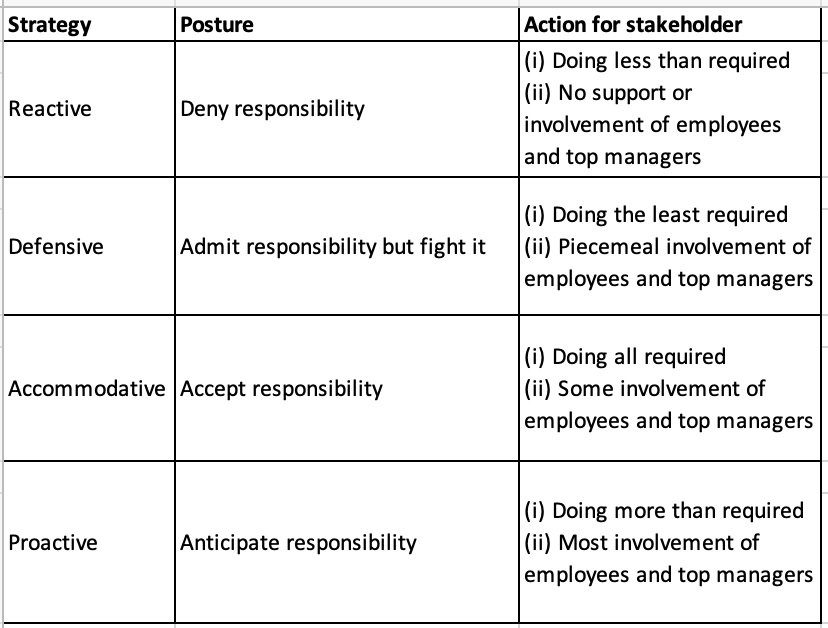}
\caption{Stakeholder Management strategy \cite{b15} }
\label{fig}
\end{figure}

Sometimes these postures of stakeholders due to the nature of the organization of the nature of data science application. Sometimes the posture varies at the department level, even in the same organization. Example in a Banking and finance business posture of the same stakeholder may vary from a Credit Risk project to a pure customer marketing project. Hence stakeholder communication and management should be strategic, and one strategy can't be generalized for all data science projects.

Another key challenge most data scientists mention is that most projects struggle to articulate the business problem well enough. Business problem understanding is not easy; it usually happens with stakeholders, business subject matter experts, and the data science team. This isn't easy because it requires a substantial amount of prior business knowledge to understand the business objective clearly. Another challenge sometimes it is hard to define a uniform business process or rule even in the same department. For example, in a credit risk modeling scenario, the objective is to identify the risky customer based on historical default customer data. But defining a default customer sometimes varies from one business SME to another business SME. Some SMEs can define a customer as a default customer if someone has not made the payment 60 days after the due date.
In some cases, SME may consider 90 days or sometimes 1st-time defaulter is not true defaulter as 1st time can be due to negligence or due to some mistake. So it's critical to understand the business problem definition clearly and once you get some level of understanding, validate that understanding in a meeting with business SME. Sometimes stakeholder comes from a mindset predictive model or machine learning a solution to any type of business problem. In some cases, it is true, but the definition of the business rule should be clear enough to make any Data Science project successful. \cite{b13} stakeholders are sometimes not sure what they want. This usually happens in business strategy projects. In this business understanding process, Data scientists should also take the opportunity to help refine the stakeholder objective in several meetings.  
But often, business understanding is not given priority because of the following reason. 
\begin{enumerate}
\item Stakeholder and Data scientist are both very excited to start working on the project
\item Data scientists cannot ask the right questions initially due to a lack of domain knowledge. 
So to solve this issue, business SME and Data Scientists should spend a good amount of time articulating the business problem clearly. 
\end{enumerate}
Another challenge in Data Science we found from this research is that the data quality. Data quality has always been an issue. Spending a good amount of time on fixing Data Quality is always difficult because effort/time to value mapping in data quality framework is always missing. Hence fixing data quality is never get prioritized. 
To understand data quality, we need to understand what data quality means. 
Data quality has three dimensions as mentioned in \cite{b20}. 
\begin{enumerate}
    \item Accuracy : Are the data free of error ?
    \item Consistency in Timeliness : Are the data up-to-date ?
    \item Completeness : Are necessary data present ?
\end{enumerate}

Reasons: Data Quality issues arise due to three main reasons. 
\begin{enumerate}
    \item Standardized Data collection process across organization margin someone's in saying about group Simpson someone's in a number of energy
    \item Error due to human or software
    \item Error in business logic to process the data
\end{enumerate}

Solution we proposed to fix data quality : 
\begin{enumerate}
    \item Better data collection standard from application level data generation
    
    \item Centralized data store : 
    An organization needs to spend time building a feature store with the help of a robust data pipeline. The objective of the feature store is to create a dataset which can be of high quality and useful for various DS projects. Ideally the team should list down all the possible DS projects and start mapping the possible features/ data which need to be created.
    
\end{enumerate}

Another key issue identified from this research is the lack of department-level collaboration, leading to low-value outcomes. Because the department which should use the product doesn't use the model as they might have other priorities to attend.  
The solution is to list down all the departments going to be affected by this outcome and department who are going to use the outcome and which department is going to be involved in hosting this solution. This need to be discussed keeping stakeholder in loop. After this department level engagement Data Science team need to share the deployment plan mention how the outcome can be deployed or reusable. 
Other issues such as change management and data governance came out of the theme are not the key bottleneck for the data science project success. However one participant mentioned that "around 30\% of the time they loose the project due to data governance and privacy limitation" - C3.
The earlier Data Science assess the data privacy and governance issues, that is better for the project. So that Data Science can have the opportunity to negotiate or can find other source which can be approved by governance.

We discussed the problems and the solution in the discussion section. In this section, we have summarized the solutions based on interviews, my knowledge, and my experience in Data Science.
Based on this research work, we would recommend there are three key areas where improvement needed
\begin{enumerate}
\item Stakeholder Management
\item Data Quality 
\item Durable and deployable outcome
\end{enumerate}
A potential data science project might fail due to the above reasons what we discussed so far. The Data Science team needs to have good stakeholder management skills to set the expectation of the stakeholder at an early stage based on the business problem understanding. Also, a data scientist should have the ability to highlight the potential risk ahead of time that might occur due to data privacy and/or data security and/or data governance restrictions. Also, Data scientists should have visibility of the risk of failure due to factors such as data quality issues and/or deployment limitations due to the complexity of the software platform and/or IT infrastructure. To create a fail-proof data quality management for the Data Science project, the Data Science team needs to spend dedicated time checking the data quality and plan to engage the data engineer to improve the data quality standard. Also, to validate the requirement with stakeholder Data Scientist need to build a deployment plan ahead and share it with the stakeholder in the early phase of the project.  By combining all, we are proposing a new Data Science methodology called HYBRID CRISP DS(Hybrid Cross Industry Standard Process for Data Science), which will take care of all the above hindrances and make Data Science project less prone to failure.


Following figure-II represents HYBRID-CRISP-DS. 

\begin{figure}[htbp]
\includegraphics[width=0.55\textwidth]{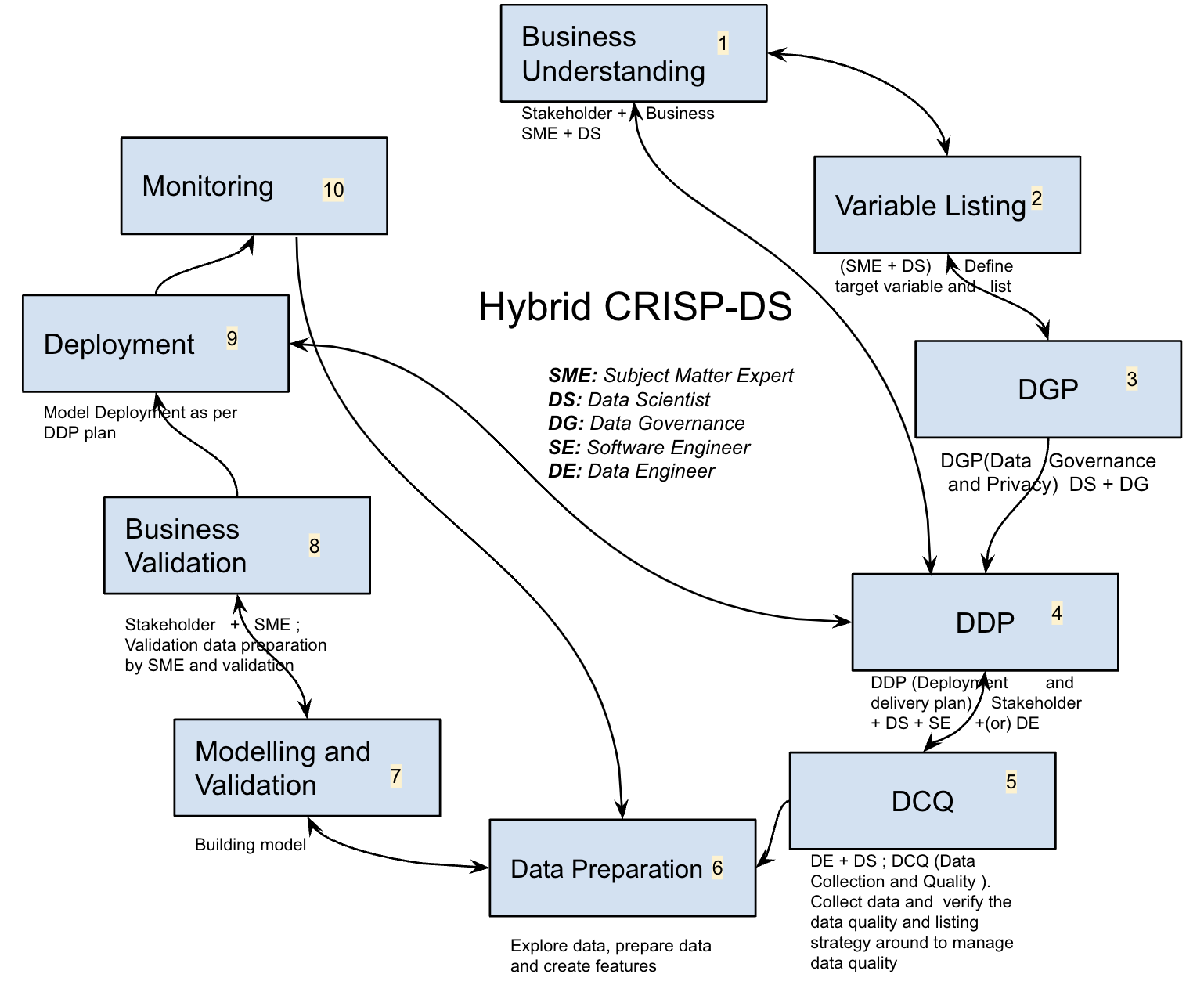}
\caption{HYBRID-CRISP-DS}
\label{fig}
\end{figure}

In this HYBRID-CRISP-DS methodology stakeholder and business SME always be in the loop during the various phases of the Data Science project. 
As mentioned in the figure-II, emphasis has been given on variable selection after consulting a business SME(subject matter expert), which is an iterative process. Soon after, that needs to be validated through DGP (Data Governance and Privacy) in collaboration with DG(Data Governance) team. Based on the outcome from DGP, proceed to the next step, DDP(deployment and delivery plan), or negotiate on variable selection for those variables rejected by DG. Based on the complexity of requirements in DDP SE(software engineer) or Data engineer, dependencies will be created. Once the deployment plan is ready, that needs to be validated by stakeholders. 
Then, the project flows through DCQ(Data collection and quality) to find the right data source and fix the quality issues where needed. After DCQ, next is to prepare the data, will follow through with model building and validation. Then the emphasis is given to outcome validation by the business user. The accuracy of a predictive model or any unsupervised model may not make sense to the business. Businesses should always test the model with live data or any validation data on which businesses can rely upon. Once the business is satisfied with the outcome, then the project moves to deployment. Once the project is deployed, a proper drift monitoring technique is implemented to monitor the drift in an automated fashion.

\section{Future work}
Although several industry experts and researchers suggested alternate approaches to CRISP-DM such as TDSP(team data science process) by Microsoft \cite{b25} SEMMA(Sampling, Exploring, Modifying, Modeling, and Assessing) by SAS \cite{b28}, each comes with the limitation. HYBRID-CRISP-DS is a new approach that came out of this research. In the future, this approach needs to be applied and can be observed where issues might occur while applying HYBRID-CRISP-DS in real-world data science projects. In this process, the Data Science project outcome never looked from the Data Science project as a software product lens. So the testing ends in step 8 of this proposed methodology after the business validates the modeling outcome. So the HYBRID-CRISP-DS approach can be researched and analyzed further to the Data Science as a product scenario, where the methodology can be slightly different with respect to business requirement, deployment, and software product testing. In addition to that, the proposed HYBRID-CRISP-DS is a series of processes. Hence this research can be further extended to understand how HYBRID-CRISP-DS process can be paralleled to make the Data Science process more efficient. 

Data Science experts participated in this interview process, not emphasizing model explainability. However, model explainability can be a big roadblock in other domains such as healthcare, critical safety environment. 
Even though this study has been conducted by interviewing experts from various domains and geography, this study can be further extended by including more professionals from a data science background.

\section*{Acknowledgment}
Special thanks to my supervisor Dr Kelly Blincoe (Senior Lecturer in Software Engineering, University of Auckland) for her expert advise, guidance and excellent support during this research. 
We acknowledged all the candidates ((Pragyansmita Nayak (C1), Confidential (C2), Mani Kanteswara Rao Garlapati (C3), Peter Devine (C4), Tural Gulmammadov (C5), Suresh Bommu (C6), Confidential (C7), Siddhant Gawsane (C8), Confidential (C9)) for their participation in this research interview.


\cleardoublepage

\end{document}